\pdfoutput=1

\documentclass[11pt]{article}

\usepackage[final]{latex/acl}

\usepackage{times}
\usepackage{latexsym}

\usepackage[T1]{fontenc}

\usepackage[utf8]{inputenc}

\usepackage{microtype}

\usepackage{inconsolata}

\usepackage{graphicx}

\usepackage{todonotes}
\usepackage{algorithm}
\usepackage[noend]{algpseudocode}
\algnewcommand\algorithmicforeach{\textbf{for each}}
\algdef{S}[FOR]{ForEach}[1]{\algorithmicforeach\ #1\ \algorithmicdo}
\usepackage{amsmath}
\usepackage{amsfonts}
\usepackage{amssymb}
\usepackage{amsmath}

\usepackage[draft]{proofing} 

\usepackage{tikz-dependency}
\usepackage{multirow}

\title{Formalising lexical and syntactic diversity for data sampling in French}

\author{Louis Estève ~~~~~~~ Manon Scholivet ~~~~~~~ Agata Savary\\
         Université Paris-Saclay, CNRS, LISN, Orsay, France \\
         {\tt firstname.lastname@universite-paris-Saclay.fr}}

\begin{document}
\maketitle
\begin{abstract}
    Diversity is 
    an important property of datasets 
    and sampling
    data for diversity is useful in 
    dataset creation. Finding the
    optimally diverse sample is expensive, we therefore present a heuristic
    significantly increasing diversity relative to random sampling.
    We also explore whether different kinds of diversity -- lexical and syntactic -- correlate, with the purpose of
    sampling for expensive syntactic diversity through inexpensive lexical diversity. 
    We find that correlations fluctuate with different datasets and versions of diversity measures. This shows that an arbitrarily chosen measure may fall short of capturing diversity-related properties of datasets.   
\end{abstract}


\section{Introduction}
\label{sec:intro}
Linguistic diversity is gaining a lot of attention in recent NLP research, for a lot of practical reasons.  
For instance, a desirable property of models is to perform comparatively well on diverse sets on languages \cite{liu-etal-2024-multilingual,gueuwou-etal-2023-jwsign}. Outputs of generative models, notably in dialog systems, are expected to be not only accurate but also diverse \citep{gao-etal-2019-jointly,alihosseini-etal-2019-jointly,xiang-etal-2024-diffusiondialog,kim-etal-2024-improving}. More diverse parameters (\emph{e.g.}, attention vectors) make a model more relevant \cite{huang-etal-2019-multi} and less sensitive to adversarial attacks \citep{yang-etal-2024-pad}. 
Diversity of human productions is seen as upper bound for the diversity of a model's output  
\citep{liu-etal-2024-prefix}. Diversity of output data being close to the one of input data is an indicator of fairness, representativeness or accurateness, \emph{e.g.}, in summarisation \cite{zhang-etal-2024-fair} and active learning \cite{xia-etal-2024-hallucination}. 
More diverse training data lead to better system performances \emph{e.g.} in 
semantic parsing  \citep{liu-zeldes-2023-cant}, dialog systems \cite{larson-etal-2019-outlier}, or question answering \cite{yadav-etal-2024-explicit}. To achieve such diverse datsets, data sampling based \emph{e.g.} on embedding outlier detection \cite{larson-etal-2019-outlier} or uncertainty-and-diversity-based active learning \cite{kim-2020-deep} can be used.

In these works, the notion of diversity is used rather loosely, or is based on measures which are selected in an ad hoc way. 
We believe that this is largely due to the unawareness of long-standing formal approaches to diversity developed in other scientific domains, notably ecology (\emph{cf.}, §\ref{sec:diversity}).

In this paper we are interested in applying ecology-inspired diversity measures, notably entropy, to the problem of data sampling. We aim at building a large corpus of French, automatically parsed for morphosyntax, with two constraints. Firstly, it should be large but manageable, \emph{i.e.} its parsing, storage and maintenance cost should not be prohibitive. Secondly,  it should still have sufficient lexical and syntactic diversity to cover long-tail phenomena.\footnote{Long-tail phenomena are important \emph{e.g.} in frame induction \cite{qasemizadeh-etal-2019-semeval}, 
zero-shot identification of multiword expressions 
\cite{ramisch-etal-2020-edition}, probing language models for rare but interesting syntactic phenomena \cite{misra2024languagemodelslearnrare,weissweiler2024hybridhumanllmcorpusconstruction}, etc. }

To this aim, we use very large raw corpora 
and we rely on formal lexical and syntactic diversity measures. We face 
 two tractability problems: (i) while lexical diversity can easily be calculated for a raw text, finding the optimally diverse subset of a very large set of texts is intractable, (ii) syntactic diversity quantification requires the 
data to be parsed in advance, which is prohibitive with very large data. 
%
In this context, we propose a data sampling heuristic which is faster than an optimal method. We address the following research questions:

\renewcommand{\theenumi}{\textbf{Q\arabic{enumi}}}
\renewcommand{\labelenumi}{\theenumi}

\begin{enumerate}
\item\label{r1} Can this method select a corpus whose diversity is significantly higher than at random?

\item\label{r2} Can lexical diversity help construct a syntactically diverse large corpus?

\end{enumerate}

The experimental results allow us to give a positive answer to \ref{r1}. Concerning \ref{r2}, we show that it is necessary not to fix one's attention on one diversity measure (here  Shannon-Weaver entropy) but to rather examine its generalisation \citep{renyi_measures_1961}. 


\section{Diversity measures}
\label{sec:diversity}
\newcommand{\removecontentenabled}[0]{1}
\newcommand{\removecontent}[1]{\ifx\removecontentenabled\0{}#1\fi}

%

{
\def\customedgeunitdistance{1.75ex}

\begin{figure*}
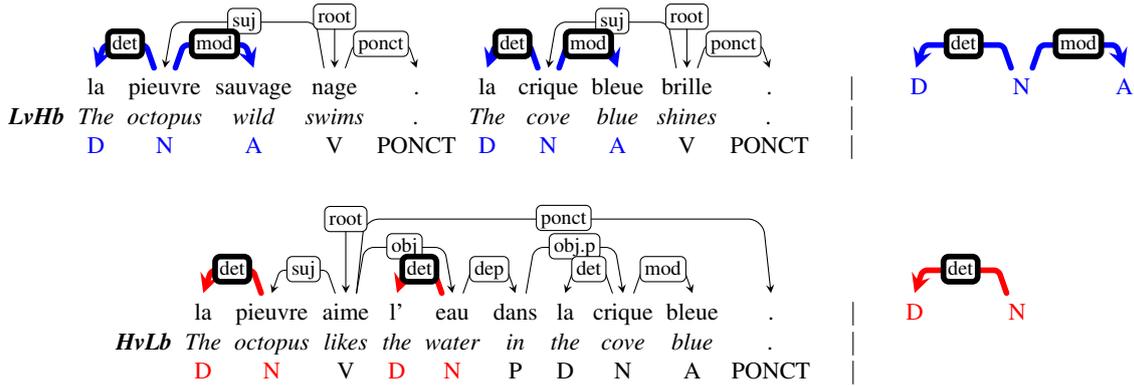

\centering
\begin{dependency}
\begin{deptext}[font=\footnotesize]
\& la \& pieuvre \& sauvage \& nage \& . \& la \& crique \& bleue \& brille \& . \&   \&  \&  $\vert$   \& \textcolor{blue}{~~~~~~~D~~~~~~} \&  \textcolor{blue}{~~~~~~N~~~~~~} \&  \textcolor{blue}{~~~~~~A~~~~~~}   \\
\emph{\textbf{LvHb}} \& \textit{The} \& \textit{octopus} \& \textit{wild} \& \textit{swims} \& \textit{.} \& \textit{The} \& \textit{cove} \& \textit{blue} \& \textit{shines} \& \textit{.} \&   \&  \&  $\vert$   \&   \&   \&   \\
\& \textcolor{blue}{D} \&  \textcolor{blue}{N} \&  \textcolor{blue}{A} \& V \& PONCT \& \textcolor{blue}{D} \&  \textcolor{blue}{N} \&  \textcolor{blue}{A} \& V \& PONCT \&    \&  \&  $\vert$   \&  \& \& \\
\\
\end{deptext}
\depedge[edge unit distance=\customedgeunitdistance]{5}{3}{suj}
\depedge[edge unit distance=\customedgeunitdistance, blue, line width=0.75mm]{3}{2}{det}
\depedge[edge unit distance=\customedgeunitdistance, blue, line width=0.75mm]{3}{4}{mod}
\depedge[edge unit distance=\customedgeunitdistance]{5}{6}{ponct}
\deproot[edge unit distance=0.75*\customedgeunitdistance]{5}{root}
\depedge[edge unit distance=\customedgeunitdistance]{10}{8}{suj}
\depedge[edge unit distance=\customedgeunitdistance, blue, line width=0.75mm]{8}{7}{det}
\depedge[edge unit distance=\customedgeunitdistance, blue, line width=0.75mm]{8}{9}{mod}
\depedge[edge unit distance=\customedgeunitdistance]{10}{11}{ponct}
\deproot[edge unit distance=0.75*\customedgeunitdistance]{10}{root}
\depedge[edge unit distance=\customedgeunitdistance, blue, line width=0.75mm]{16}{15}{det}
\depedge[edge unit distance=\customedgeunitdistance, blue, line width=0.75mm]{16}{17}{mod}
\end{dependency}
\begin{dependency}
\begin{deptext}[font=\footnotesize]
\& la \& pieuvre \& aime \& l' \& eau \& dans \& la \& crique \& bleue \& . \&     \&  \&  $\vert$   \&  \textcolor{red}{~~~~~~D~~~~~~} \&  \textcolor{red}{~~~~~~N~~~~~~}   \\
\emph{\textbf{HvLb}} \& \textit{The} \& \textit{octopus} \& \textit{likes} \& \textit{the} \& \textit{water} \& \textit{in} \& \textit{the} \& \textit{cove} \& \textit{blue} \& \textit{.} \&    \&  \&  $\vert$   \&   \&   \\
\& \textcolor{red}{D} \&  \textcolor{red}{N} \& V \& \textcolor{red}{D} \&  \textcolor{red}{N} \& P \& D \& N \& A \& PONCT \&   \&  \&  $\vert$   \&  \&  \\
\\
\end{deptext}
\depedge[edge unit distance=\customedgeunitdistance]{4}{3}{suj}
\depedge[edge unit distance=\customedgeunitdistance]{4}{6}{obj}
\depedge[edge unit distance=\customedgeunitdistance]{6}{7}{dep}
\depedge[edge unit distance=0.81ex]{4}{11}{ponct}
\depedge[edge unit distance=\customedgeunitdistance, red, line width=0.75mm]{3}{2}{det}
\depedge[edge unit distance=\customedgeunitdistance, red, line width=0.75mm]{6}{5}{det}
\depedge[edge unit distance=\customedgeunitdistance]{7}{9}{obj.p} 
\depedge[edge unit distance=\customedgeunitdistance]{9}{8}{det}
\depedge[edge unit distance=\customedgeunitdistance]{9}{10}{mod}
\deproot[edge unit distance=\customedgeunitdistance]{4}{root}
\depedge[edge unit distance=\customedgeunitdistance, red, line width=0.75mm]{16}{15}{det}
\end{dependency}
\caption{Two toy datasets with sample syntactic categories (on the right) and elements (inside the sentences).}\label{fig:trees-as-categories}
\end{figure*}

}

Formal approaches to diversity \citep{morales_measuring_2020,chao_unifying_2014} apportion \textbf{elements} into \textbf{categories}. For instance in ecology -- a field with a long history of formal diversity --, species are categories and individuals are elements.

We apply this principle to NLP datasets.
For \textbf{lexical diversity}, categories are unique 
token forms (henceforth: \emph{forms}) and elements are their occurrences. 
For instance in Fig.~\ref{fig:trees-as-categories}, datasets \emph{\textbf{LvHb}} (Low Variety High Balance) and \emph{\textbf{HvLb}} (High Variety Low Balance) have $m=10$ elements each, and $n=8$ and $n=9$ categories, respectively.
For \textbf{syntactic diversity}, categories are complete syntactic subtrees containing only POS labels and dependency relations, and elements are all instances of these subtrees. For instance, in Fig.~\ref{fig:trees-as-categories}, \emph{\textbf{LvHb}} and \emph{\textbf{HvLb}} have 5 and 8 categories, respectively (\emph{cf.}, Appendix~\ref{tab:extracted-trees}), including the ones shown on the right, having two elements highlighted in blue and red. 

Given $n$ categories and $m$ elements, diversity is measured along 3 dimensions: \textbf{variety}, \textbf{balance}, and
\textbf{disparity} \citep{morales_measuring_2020}. We here only discuss variety and balance.

\textbf{Variety} tackles the number of
categories:
a (pure) variety function
is monotonic to $n$ and its most basic form is richness, \emph{i.e.}, just $n$. 
Thus, a habitat with $n$ species is more varied than with $n-1$, and \emph{\textbf{HvLb}} with $9$ forms and $7$ subtress is lexically and syntactically more varied 
than \emph{\textbf{LvHb}} with $8$ forms and $5$ subtrees.

\textbf{Balance} tackles 
the distribution of elements in categories.
A (pure) balance function reaches its maximum
 when all categories are equiprobable.
Thus, a habitat with 50 octopuses and 50 squids is more balanced than one with 60 and 40. \emph{\textbf{LvHb}} with 2 elements in each subtree is syntactically more balanced than \emph{\textbf{HvLb}} with subtrees having more elements than others.

A large number of measures was proposed for variety and balance in the past \citep{hill_diversity_1973,smith_consumers_1996} and many are hybrids between the two. For conciseness, we restrict our study to entropies, as they encompass many diversity functions and have a strong
background.
Consider the non-zero probability distribution of categories
$\Delta = \left\{p_1, ..., p_n\right\}$.
\citet{shannon_mathematical_1949} entropy $H$ may thus be computed as (here $b = e$): 
\begin{equation}
H\left(\Delta\right) = -\sum\limits_{i=1}^{n} p_i \log_b\left(p_i\right)
\end{equation}
When $\forall p_i \in \Delta, p_i = \frac{1}{n}$, $H$ reaches its maximum $\log_b\left(n\right)$. 
It is thus hybrid of
variety and balance.
\citep{renyi_measures_1961} generalizes entropy as $H_\alpha$, which is pure variety
when $\alpha=0$. With growing $\alpha$, $H_\alpha$ accounts more for balance and less for variety:
\begin{equation}
H_1=H; H_{\alpha \neq 1}\left(\Delta\right) = \frac{1}{1 - \alpha} \log_b\left(\sum\limits_{i=1}^{n}p_i^\alpha\right)
\end{equation}
Using data of our toy examples, we find:

\begin{tabular}{|c|c|c|c|c|} \hline
    &   \multicolumn{2}{|c|}{\emph{LvHb}}   &   \multicolumn{2}{|c|}{\emph{HvLb}}  \\ \hline
    Dataset &   Lex. &   Syn.   &   Lex. &   Syn.   \\ \hline
    $H_0$   &   2.079 & 1.609 & 2.197 & 2.079                   \\ \hline
    $H_1$   &   2.025 & 1.609 & 2.163 & 1.973                   \\ \hline
    $H_2$   &   1.966 & 1.609 & 2.120 & 1.832                   \\ \hline
\end{tabular}



\section{Source data}
\label{sec:data}
We wish to build a large syntactically parsed French corpus, with sufficient lexical and syntactic diversity to serve lexicon induction and morphosyntactically-based studies. We use the FTB-dep annotation schema \cite{candito-etal-2010-statistical} for its ability to finely represent French syntax.

Our first source, now called \BASE{$BASE$}, is BigScience \cite{NEURIPS2022_ce9e92e3}, and more precisely its three subcorpora: Europarl, French part of the United Nations Corpus and Wikipedia\footnote{Wikisource was also considered, but had many issues.}, for a total of 833K documents and 1.5 billion tokens. Their advantages are large sizes, clearly identified sources and genres (parliamentary debates and encyclopedia entries), few multilingual texts, and clear licenses inspired by openness and fairness.

A major disadvantage of \BASE{$BASE$} is not to be sufficiently diverse as far as genres are concerned, which likely influences lexical and syntactic diversity. Therefore, we use HPLT \cite{de2024new}, a massive multilingual dataset provided by Internet Archive and CommonCrawl. In the \emph{cleaned} version of HPLT  v1.2, texts were partly filtered for dubious sources (pornographic, racist, \emph{etc}) and noisy paragraphs. They were also divided into languages by majority vote over a number of language predictors. We use the texts assigned to French, for a total of 99.59M documents and 122.88B tokens. Less clean than BigScience (due to imperfect filtering and language prediction), this dataset
covers a wide range of fields and will be useful to increase the diversity of
\BASE{$BASE$}.

\section{Diversity-driven data selection}
\label{sec:data-selection}
As HPLT contains too much text, we 
select only a number of tokens similar to BigScience (\emph{i.e.}, around 1.5 billion). To this aim, we use a data augmentation process driven by lexical diversity, which we define here as entropy (\emph{cf.}, §\ref{sec:diversity}), where elements are tokens and categories are unique forms.

We start by randomly selecting around 6 billion documents from HPLT,
to keep
computation time reasonable. Henceforth, this subset is called \HPLT{HPLT$_{small}$}. 
We then apply Algorithm~\ref{alg:heuristic_sample}.
%
Its objective is to extend a dataset with a restricted number of new documents, while maximising lexical diversity. Precisely, given an initial dataset \BASE{$I$} (here \BASE{$BASE$}), a new dataset \HPLT{$N$} (here \HPLT{HPLT$_{small}$}), and a target size $S$ (here 3.1 billion), we select documents from \HPLT{$N$} which, added to \BASE{$I$}, increase \BASE{$I$}'s size to $S$ 
while maximising lexical diversity. Documents from \HPLT{$N$} selected this way to be added to \BASE{$I$} are called \DIVERSE{$N_{diverse}$} (here \DIVERSE{HPLT$_{diverse}$}), and the final corpus, containing \BASE{$I$} and \DIVERSE{$N_{diverse}$} is called \TOTAL{$TOTAL$}.

The optimal choice of \DIVERSE{$N_{diverse}$} would require a very high computational cost.
Consider
a dataset as a
set of undividable blocks $\mathbb{A}$.
(\emph{e.g.}, a set of sentences
or a
set of texts).
The question is: which
subset $\mathbb{B} \subseteq \mathbb{A}$ maximises $H_\alpha$? An exhaustive search of
$\mathcal{P}\left(\mathbb{A}\right)$ -- the power set of $\mathbb{A}$, the set of all possible subsets of $\mathbb{A}$ -- would take $O\left(2^{\left\vert \mathbb{A}
\right\vert}\right)$, which is not tractable.
Therefore, we use a heuristic to approximate it.

We start with the working corpus $W$ equal to \BASE{$I$} (l.~\ref{alg:init-W}).
In the internal loop (l.~\ref{alg:foreach-document}-\ref{alg:foreach-document-end}), we consider one candidate document $n$ from \HPLT{$N$} at a time. We filter and normalise it (l.~\ref{alg:norm}) to avoid artificial increase in diversity.\footnote{(Telephone) numbers, HTML and XML tags, URLs, file paths, emoticons, series of punctuations, phonetic characters, series of alphanumerical tokens, and characters outside of the French range are represented by a unique token for each category, \emph{e.g.}, \texttt{[NUMBER]}.
}
We check if, added to $W$, $n$ increases entropy (l.~\ref{alg:test-increases-entropy}). If so, we check if this increase in entropy is higher than for a previously found document $d$ (l.~\ref{alg:test-increases-entropy-more-than-current-best}). If so,
$d$ becomes $n$ (l.~\ref{alg:replace-d-with-n}). 
Note that, the next optimal document $d$ to add is not chosen from the whole corpus \HPLT{$N$} but, for tractability, we stop when we found "enough" documents increasing entropy (l.~\ref{alg:increase-exhaustivity-counter} and \ref{alg:test-fulfilled-current-exhaustivity-level}). Then we pick the best of them $d$ and add it to the working corpus $W$ (l.~\ref{alg:append-to-corpus}). 

Variable $e$, for exhaustivity of search, tells us how many documents we have to look at before we pick the optimal one to append to $W$. If $e$ is too high, we might not reach the intended size $S$. Therefore, the array $E$ gives several exhaustivity values in decreasing order. If $S$ has not been reached, we become less exhaustive, \emph{i.e.}, we go to the next $e$.

We return $W$ (which then becomes \TOTAL{$TOTAL$}) as soon as it has exceeded $S$ (l.~\ref{alg:test-fulfilled-maximum-size}), or when all exhaustivity levels have been considered. 

\begin{algorithm}
\small
\begin{algorithmic}[1]
    \Require \BASE{$I$}, an initial dataset
     \Require \HPLT{$N$}, another dataset to select from 
    \Require $E$, a decreasing array of exhaustivity search parameters (positive integers)
    \Require $S$, maximum size of resulting dataset 
    \Ensure \BASE{$I$} increased with fragments of \HPLT{$N$} (while maximising entropy)
    \State $W \gets \BASE{I}$, the working dataset \label{alg:init-W}
    \ForEach {$e \in E$}, for each exhaustivity level \label{alg:foreach-exhaustivity-level}
        \State $f \gets 0$, counter for current exhaustivity
        \State $d \gets \emptyset$, document to append
        \ForEach {$n \in N$}, for each document \label{alg:foreach-document}
            \State $n \gets \text{normalised}\left(n\right)$ \label{alg:norm}
            \If{$H\left(W \cup n\right) > H\left(W\right)$} \label{alg:test-increases-entropy}
                \State $f \gets f + 1$ \label{alg:increase-exhaustivity-counter}
                \If{$H\left(W \cup n\right) > H\left(W \cup d\right)$} \label{alg:test-increases-entropy-more-than-current-best}
                    \State $d \gets n$ \label{alg:replace-d-with-n}
                \EndIf
                \If{$f = e$} \label{alg:test-fulfilled-current-exhaustivity-level}
                    \State $W \gets W \cup d$ \label{alg:append-to-corpus}
                    \State $f \gets 0$
                    \State $d \gets \emptyset$
                    \If{$\left\vert W \right\vert > S$} \label{alg:test-fulfilled-maximum-size}
                        \Return $W$
                    \EndIf
                \EndIf
            \EndIf
        \EndFor \label{alg:foreach-document-end}
    \EndFor \label{alg:foreach-exhaustivity-level-end}    
\Return $W$\label{alg:return}
    
\end{algorithmic}
\caption{Algorithm to heuristically sample data while maximising entropy.\label{alg:heuristic_sample}}
\end{algorithm}


%

\section{Diversity evaluation}
\label{sec:evaluation}
The initial lexical $H$ of \BASE{$BASE$} is 7.02 for 1,538,617,909 tokens.  Applying Algorithm~\ref{alg:heuristic_sample}, $TOTAL$ 
obtains an entropy $H_{diverse}$ of 7.98. While 0.96 may appear as a small increase, it is in fact large due to the logarithmic nature of entropy.
%
To verify the effectiveness of 
Algorithm~\ref{alg:heuristic_sample} in selecting a diverse corpus, 
we will answer the two research questions defined in §\ref{sec:intro}. 

\paragraph{Q1} 
To test \ref{r1}, 
in
addition to \DIVERSE{$N_{diverse}$}, we construct \RANDOM{$N_{random}$} by randomly selecting sentences from \HPLT{$HPLT_{small}$} until reaching $S$. We repeat this with 20 different seeds.
We test whether
entropy
for these 20 random samples follows a normal distribution $H_{\textit{random}} \sim \mathcal{N}\left(x;\mu,\sigma\right)$.
We consider the null hypothesis $h_0$ to be \emph{$H_{\textit{random}}$ follows a normal distribution}, and the alternative hypothesis $h_1$ to be \emph{it does not follow a normal distribution}.\footnote{See Python's \texttt{scipy.stats.normaltest}.}
As it is not significant ($p \approx 0.26$), it is likely we have a normal distribution.

For this normal distribution, $\mu$ $\approx 7.656$ and $\sigma$ $\approx 9.027\text{e}-4$, and the value of $H_{diverse}$ is away from $\mu$ by $\approx 347\sigma$. The $p$-value is $\approx 0$.\footnote{$p$ is lower than the precision
of
a 64-bit double.} Thus, \DIVERSE{$N_{diverse}$} is highly significantly more diverse than at random, which confirms the effectiveness of Algorithm~\ref{alg:heuristic_sample}.

\paragraph{Q2} 
Checking the correlation between lexical and syntactic diversity requires the corpus to be parsed, which is prohibitive even for \HPLT{HPLT$_{small}$}.
Since \BASE{$BASE$} has already been parsed, we use it for the correlation estimation.
We split it into 705 blocks of 100K sentences. Lexical and syntactic $H_\alpha$ with $\alpha$ ranging from $0$ to $5$ are calculated for each block.
Spearman and Pearson correlations 
are calculated at each level of $\alpha$.



\begin{figure}
    \includegraphics[width=\columnwidth]{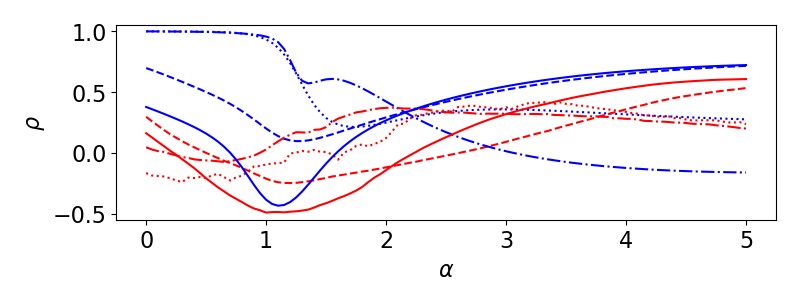}
    \caption{Correlation between lexical and syntactic $H_\alpha$, according to $\alpha$. Europarl (dotted), UN corpus (dashed), Wikipedia (dash-dotted), and the union of all three (solid). Blue for Pearson, red for  Spearman.\label{fig:correlations}}
\end{figure}

We find that the Correlation scores between Lexical and Syntactic Diversities (CLSD) cannot easily be exploited (\emph{cf.}, Fig.~\ref{fig:correlations}). 
Firstly, the choice of $\alpha$ highly impacts CLSD. The precise choice of $\alpha$ (\emph{i.e.}, whether considering more variety or more balance) must thus be thought through.
Secondly, across datasets and $\alpha$ values, CLSD is not constant. Approximating syntactic through lexical diversity would therefore require some amount of parsing to see if for the studied dataset CLSD are usable. 
Thirdly, within a specific genre, CLSD do not seem consistent: we see little 
resemblance between Europarl and UN corpus, despite both of them being parliamentary debates.
Finally, the union of datasets (\emph{cf.}, solid curves) 
shows special properties: while 
Pearson correlation is positive or slightly below $0$ for subcorpora (Europarl, UN corpus, and Wikipedia), the union dips at $-0.43$ near $\alpha=1$.

These results might be interpreted by non-compositionalty of the diversity calculus. Given two datasets $D1$ and $D2$, the diversity of their union depends on their similarity. For instance, the variety of the union depends on if $D1$ and $D2$ have disjoint categories or not. The balance of the union can be high when $D1$ and $D2$ are unbalanced but share categories with inverse distribution patterns. More insight is needed to exploit these results.

These results do not provide a definitive conclusion for \ref{r2}. The use of lexical diversity did not increase syntactic diversity, but there are still research directions to be explored in this area.

\section{Conclusions and future work}
\label{sec:conclusions}




We proposed an algorithm for diversity-driven data sampling which is tractable compared to optimal solution and still significantly increases the lexical diversity of a dataset.
We showed that correlation between lexical and syntactic diversities is not reliable enough for syntactic diversity to be approximated by lexical diversity.
The algorithm may be use to sample for high syntactic diversity, but more insights are needed to reduce annotation cost.

\section{Limitations}
\label{sec:limitations}
The block size in the second experiment is arbitrary, as such we cannot ensure that another value would have yielded the same results.
Our experiments are limited to French.
The way the algorithm is coded favours sentences at the beginning of files, and longer documents. 
The syntax used to compute the syntactic diversity is predicted and as such contains errors.
Some rare phenomena might especially be badly predicted, which may impact the diversity scores.

\section{Ethical statement}
\label{sec:ethical-statement}
The algorithm we presented in this article is agnostic of the data. 
As such, when given a "clean" \BASE{$BASE$} (\emph{i.e.}, that does
not contain undesirable content), it tends to select previously unseen data, since such data often increase diversity scores.
The selected data may have inappropriate content (\emph{e.g.}, pornographic or racist data), 
even after filtering for data sources, as filters are often imperfect.

The authors have no known conflict of interest with the authors of source data.


\bibliography{custom.bib,diversity.bib}

\appendix

\section{Appendix}
\def\customedgeunitdistance{1.75ex}

\begin{table}[ht]
    \centering
    \small
    \begin{tabular}{|c||c|c||p{0.7\columnwidth}|c|}
        \hline
        Dataset    &   \multicolumn{2}{|c||}{Lexical} &   \multicolumn{2}{|c|}{Syntactic} \\ \hline
            &   Category & Nb Elem &   Category & Nb Elem \\ \cline{2-5}
        \multirow{8}{*}{\emph{LvHb}}     
        &    .   &   2  &   \begin{minipage}[c][1.5cm][c]{1cm}
          
          \begin{dependency}
            \begin{deptext}
              D~ \& ~N~ \& ~A~ \& ~V \& PONCT\\
            \end{deptext}
            \depedge[edge unit distance=\customedgeunitdistance]{2}{1}{det}
            \depedge[edge unit distance=\customedgeunitdistance]{2}{3}{mod}
            \depedge[edge unit distance=1.25*\customedgeunitdistance]{4}{2}{suj}
            \depedge[edge unit distance=\customedgeunitdistance]{4}{5}{ponct}
          \end{dependency}
        \end{minipage}  &  2   \\ \cline{2-5}
        &    la  &   2   & \begin{minipage}[c][1.25cm][c]{1cm}
          \begin{dependency}
            \begin{deptext}
              D \& N \& A \\
            \end{deptext}
            \depedge[edge unit distance=\customedgeunitdistance]{2}{1}{det}
            \depedge[edge unit distance=\customedgeunitdistance]{2}{3}{mod}
          \end{dependency}
        \end{minipage} & 2 \\ \cline{2-5}
        &    bleue   &   1  &  D & 2 \\ \cline{2-5}
        &    brille  &   1  &  A & 2 \\ \cline{2-5}
        &    crique  &   1  & PONCT  & 2  \\ \cline{2-5}
        &    nage    &   1  &  &     \\ \cline{2-5}
        &    pieuvre &   1  &  &    \\ \cline{2-5}
        &    sauvage &   1  &  &    \\ 
        \hline \hline
        \multirow{24}{*}{\emph{HvLb}} 
        &    la   &   2  &   \begin{minipage}[c][2cm][c]{1cm}
          \begin{dependency}
            \begin{deptext}
              D~ \&  N~ \& V~ \& D~ \&  N~ \& P~ \& D~ \& N~ \& A \& PONCT \& \\
            \end{deptext}
            \depedge[edge unit distance=\customedgeunitdistance]{3}{2}{suj}
            \depedge[edge unit distance=1.25*\customedgeunitdistance]{3}{5}{obj}
            \depedge[edge unit distance=\customedgeunitdistance]{5}{6}{dep}
            \depedge[edge unit distance=0.6*\customedgeunitdistance]{3}{10}{ponct}
            \depedge[edge unit distance=\customedgeunitdistance]{2}{1}{det}
            \depedge[edge unit distance=\customedgeunitdistance]{5}{4}{det}
            \depedge[edge unit distance=1.25*\customedgeunitdistance]{6}{8}{obj.p} 
            \depedge[edge unit distance=\customedgeunitdistance]{8}{7}{det}
            \depedge[edge unit distance=\customedgeunitdistance]{8}{9}{mod}
          \end{dependency}
        \end{minipage} & 1 \\ \cline{2-5}
        &    .  &   1  &     \begin{minipage}[c][1.5cm][c]{1cm}
          \begin{dependency}
            \begin{deptext}
                 D~\&  N~ \& P~ \& D~ \& N~ \& A~  \\
            \end{deptext}
            \depedge[edge unit distance=\customedgeunitdistance]{2}{1}{det}
            \depedge[edge unit distance=\customedgeunitdistance]{2}{3}{dep}
            \depedge[edge unit distance=\customedgeunitdistance]{5}{4}{det}
            \depedge[edge unit distance=1.25*\customedgeunitdistance]{3}{5}{obj.p} 
            \depedge[edge unit distance=\customedgeunitdistance]{5}{6}{mod}
          \end{dependency}
        \end{minipage} & 1   \\ \cline{2-5}
        &    aime   &   1   &   \begin{minipage}[c][1.5cm][c]{1cm}
          \begin{dependency}
            \begin{deptext}
              P~ \& D~ \& N~ \& A~  \\
            \end{deptext}
            \depedge[edge unit distance=\customedgeunitdistance]{3}{2}{det}
            \depedge[edge unit distance=1.25*\customedgeunitdistance]{1}{3}{obj.p} 
            \depedge[edge unit distance=\customedgeunitdistance]{3}{2}{det}
            \depedge[edge unit distance=\customedgeunitdistance]{3}{4}{mod}
          \end{dependency}
        \end{minipage} & 1    \\ \cline{2-5}
        &    bleue  &   1   &   \begin{minipage}[c][1.25cm][c]{1cm}
          \begin{dependency}
            \begin{deptext}
              D \& N \& A \\
            \end{deptext}
            \depedge[edge unit distance=\customedgeunitdistance]{2}{1}{det}
            \depedge[edge unit distance=\customedgeunitdistance]{2}{3}{mod}
          \end{dependency}
        \end{minipage} & 1 \\ \cline{2-5}
        &    crique  &   1  &    \begin{minipage}[c][1.25cm][c]{1cm}
          \begin{dependency}
            \begin{deptext}
              D \& N \\
            \end{deptext}
            \depedge[edge unit distance=\customedgeunitdistance]{2}{1}{det}
          \end{dependency}
        \end{minipage} & 1  \\ \cline{2-5}
        &    dans    &   1  &  D & 3  \\ \cline{2-5}
        &    eau &   1  &  A  & 1 \\ \cline{2-5}
        &    l' &   1   & PONCT & 1 \\ \cline{2-5}
        &    pieuvre &   1   &  & \\ \hline
        
    \end{tabular}
    \caption{Extracted categories per dataset in Figure \ref{fig:trees-as-categories}.\label{tab:extracted-trees}}
\end{table}




\end{document}